\documentclass[10pt,twocolumn,letterpaper]{article}

\usepackage{cvpr}
\usepackage{times}
\usepackage{epsfig}
\usepackage{graphicx}
\usepackage{amsmath}
\usepackage{amssymb}

\usepackage{float}

\usepackage{booktabs}

\usepackage[pagebackref=true,breaklinks=true,letterpaper=true,colorlinks,bookmarks=false]{hyperref}

\cvprfinalcopy % *** Uncomment this line for the final submission

\def\cvprPaperID{****} % *** Enter the CVPR Paper ID here

\ifcvprfinal\fi
\begin{document}

%%%%%%%%% TITLE
\title{Beauty Learning and Counterfactual Inference}

\author{Tao Li\\
Department of Computer Science\\
Purdue University\\
{\tt\small taoli@purdue.edu}
% \and
}

\maketitle
%\thispagestyle{empty}

%%%%%%%%% ABSTRACT
\begin{abstract}
This work showcases a new approach for causal discovery by leveraging user experiments and recent advances in photo-realistic image editing, demonstrating a potential of identifying causal factors and understanding complex systems counterfactually.
We introduce the beauty learning problem as an example, which has been discussed metaphysically for centuries and recently been proved exists, is quantifiable, and can be learned by deep models in our paper~\cite{liu2019understanding}, where we utilize a natural image generator coupled with user studies to infer causal effects from facial semantics to beauty outcomes, the results of which also align with existing empirical studies.
We expect the proposed framework for a broader application in causal inference.
\end{abstract}

\section{Introduction}\label{sec:introduction}
Recent advances in deep learning provide the capability of capturing complex data distributions and has promoted many successful applications in various domains.
However, these models are mostly black boxes and deficit in explainability, demanding a timely study of ``distilling knowledge from deep networks''~\cite{chen2018explaining}. Moreover, deep models fail to understand causal relationships behind and struggle to help answer questions such as ``What if?'' and ``Why?''~\cite{pearl2018seven}. It is of great interests to not only enable automated systems to learn association rules from complicated and heterogenous data, but also be able to infer causal relations and reason counterfactually.
Here, we introduce a new pipeline for causal discovery, incorporating user experiments with recent advances in image editing.
In particular, we investigate the beauty learning problem, given that facial attractiveness has profound influences on multiple aspects of human society and yet is not well studied from a quantitative perspective.
In the rest of the paper, Section \ref{sec:literature} introduces necessary backgrounds in causal inference and image editing. Section \ref{sec:method} formalizes the problem and highlights our approach, showcasing that photo-realistic image editing along with user studies can be a powerful tool for counterfactual reasoning.
We leave more details and discussions in the full paper~\cite{liu2019understanding}.

\section{Preliminary}\label{sec:literature}
\paragraph*{Pearl's Causal Hierarchy}
\textit{Judea Pearl}~\cite{pearl2009causality,pearl2009causal} introduced a causal hierarchy for classification of causal information, providing a theoretical foundation for analysis of causality. The causal hierarchy encompasses three levels:
(i) association; (ii) intervention; and (iii) counterfactuals.
Table \ref{tab:ladder} below highlights the causal ladder with examples.
\begin{table}[h]
\begin{center}
\resizebox{\columnwidth}{!}{
\begin{tabular}{@{}llll@{}}
\toprule
Level                                                                           & Typical Activities                                               & Typical Questions                                                                                              & Examples                                                                                                                                                                                          \\ \midrule
\begin{tabular}[c]{@{}l@{}}1. Association\\     $P(y | x)$\end{tabular}         & Seeing                                                           & \begin{tabular}[c]{@{}l@{}}What is?\\ How would seeing $X$\\ change my belief in $Y$?\end{tabular}            & \begin{tabular}[c]{@{}l@{}}What does a symptom tell\\ me about a disease?\\ What does a survey tell us\\ about the election results?\end{tabular}                                                 \\ \midrule
\begin{tabular}[c]{@{}l@{}}2. Intervention\\     $P(y | do(x), z)$\end{tabular} & Doing                                                            & \begin{tabular}[c]{@{}l@{}}What if?\\ What if I do $X$?\end{tabular}                                          & \begin{tabular}[c]{@{}l@{}}What if I take aspirin, will\\ my headache be cured?\\ What if we ban cigarettes?\end{tabular}                                                                         \\ \midrule
\begin{tabular}[c]{@{}l@{}}3. Counterfactuals\\ $P(y_x | x', y')$\end{tabular}   & \begin{tabular}[c]{@{}l@{}}Imaging,\\ Retrospection\end{tabular} & \begin{tabular}[c]{@{}l@{}}Why?\\ Was it $X$ that caused $Y$?\\ What if I had acted differently?\end{tabular} & \begin{tabular}[c]{@{}l@{}}Was it the aspirin that\\ stopped my headache?\\ Would Kennedy be alive\\ had Oswald not shot him?\\ What if I had not been\\ smoking the past two years?\end{tabular} \\ \bottomrule
\end{tabular}
}
\end{center}
\caption{Pearl's ladder of causation~\cite{pearl2009causality,pearl2009causal} and examples.}
\label{tab:ladder}
\end{table}

\paragraph*{Structural Causal Model}
Structural Causal Model (SCM)~\cite{pearl2009causality} lies in the key of structural causal inference.
It is defined as an ordered tuple $\langle \mathbf{U}, \mathbf{V}, \mathcal{F}, \mathbf{P(u)} \rangle$, where $\mathbf{U}$ is a set of exogenous variables determined by factors outside the model, $\mathbf{V}$ is a set of endogenous variables determined by factors inside the model, $\mathcal{F}$ is a set of functions that express the structures of the model, and $\mathbf{P(u)}$ is the distribution of $\mathbf{U}$. A diagram that captures the relationships among these variables is called a causal diagram (or graph $G$).
With these notations, a causal effect from $X$ to $Y$ can be formally defined as
\begin{equation}
P(y|do(x))
\end{equation}
and the effect is said to be identifiable if $P(y|do(x))$ can be expressed and calculated using a combination and derivatives of  $\mathbf{P(u)}$.

\paragraph*{Photo-Realistic Image Editing}
Image editing, as an important topic in computer vision, has been greatly promoted by recent advances in Generative Adversarial Network (GAN), which a system of two neural networks that contests with each other under a zero-sum game setting. It was first introduced by~\cite{goodfellow2014generative} in 2014. Since then, great progresses in both theory and practice have been made, to name a few: DCGAN~\cite{radford2015unsupervised}, CycleGAN~\cite{zhu2017unpaired}, and AnonymousNet~\cite{li2019anonymousnet}.
Recently, \textit{Karras et al.}~\cite{karras2018style} proposed a style-based generator architecture (StyleGAN) that is able to synthesize photo-realistic non-existent facial images which can barely be distinguished from real ones by human eyes.

\section{The Example of Beauty Learning}\label{sec:method}
What is beauty? It has been debated by philosophers and psychologists for centuries. The answers range from symmetry~\cite{thornhill1994human}
and averageness~\cite{galton1878composite} to personality~\cite{little2006good}
and sexual dimorphism~\cite{penton2004high}.
The proverb \textit{beauty is in the eye of the beholder} implies that perception of beauty is subjective and stems from various cultural and social settings.

\subsection{Causation in Beauty Learning}
\begin{figure}[h]
\begin{center}
\includegraphics[width=0.6\linewidth]{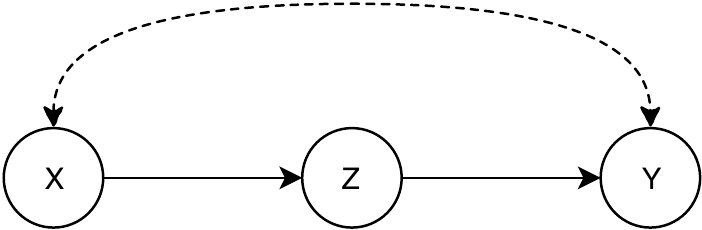}
\end{center}
\caption{Causal diagram of the SCM.
% where set $\mathbf{X}$ is latent representations of the deep nets, set $\mathbf{Z}$ represents facial attributes, and set $\mathbf{Y}$ is the visual outcomes. The model is also influenced by unobserved variables $\mathbf{U}$ as ``shown'' by the dash line.
}
\label{fig:diagram}
\end{figure}

Despite extensive discussions, many causal questions have never been well answered: ``can a machine learn beauty semantics?'' (association), ``will changing a facial attribute impact attractiveness?'' (intervention), and ``why does this facial attribute matter?'' (counterfactuals). Figure~\ref{fig:diagram} provides a structural causal diagram that formalizes the question that whether attribute $\mathbf{X}$ has a causal effect on beauty score $\mathbf{Y}$ in the present of unobserved variable $\mathbf{U}$.

\subsection{Beauty Learning from Data}
\begin{figure}[H]
\begin{center}
\includegraphics[width=0.11\textwidth]{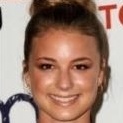}
\includegraphics[width=0.11\textwidth]{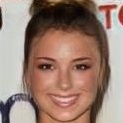}
~
\includegraphics[width=0.11\textwidth]{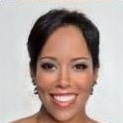}
\includegraphics[width=0.11\textwidth]{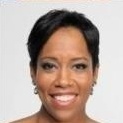}
\\
\includegraphics[width=0.11\textwidth]{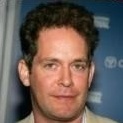}
\includegraphics[width=0.11\textwidth]{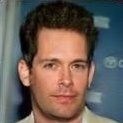}
~
\includegraphics[width=0.11\textwidth]{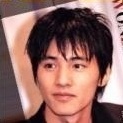}
\includegraphics[width=0.11\textwidth]{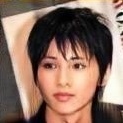}
\end{center}
\caption{
In each pair, left is the original image; right is the translated. Which one (left or right) do you think is more attractive?
}
\label{fig:results}
\end{figure}

Previous definitions of beauty are mostly subjective and metaphysical, and deficit in accuracy, generality, and scalability.
To overcome these issues, in~\cite{liu2019understanding} we present a data-driven study in mining beauty semantics of facial attributes, in an effort to objectively construct descriptions of beauty in a quantitative manner.
We first deploy a deep Convolutional Neural Network (CNN)~\cite{szegedy2015going} trained by the CelebA dataset~\cite{liu2015deep} to extract facial attributes.
Then we investigate correlations between these facial attributes and attractiveness on two large-scale datasets with labelled beauty scores (Beauty 799 dataset~\cite{zhang2016new} and US 10K dataset~\cite{bainbridge2013intrinsic}) and accordingly select key attributes for beauty enhancements supported by statistical tests (e.g., \texttt{small nose}, \texttt{high cheekbones}, and \texttt{femininity}).
We further leverage a Generative Adversarial Network (GAN)~\cite{choi2018stargan} to translate these high-level representations from original images to beauty-enhanced alternatives. Figure~\ref{fig:results} illustrates some results.

\subsection{User Study}
\begin{figure}[H]
\begin{center}
\includegraphics[width=0.95\linewidth]{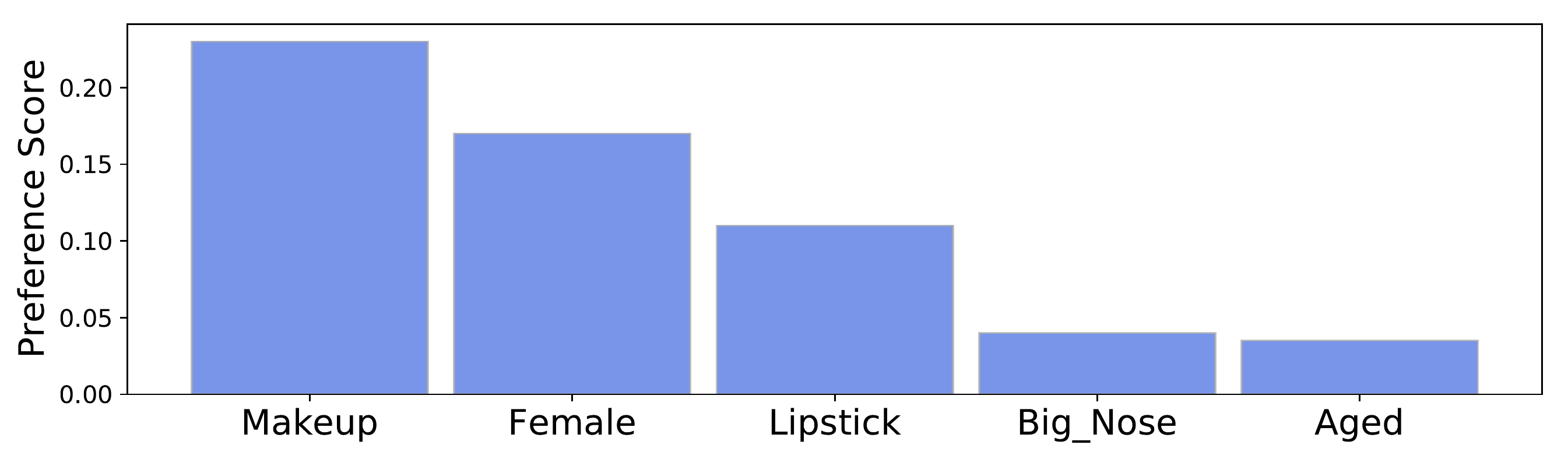}
\end{center}
\caption{Results from the user experiment.
% We perform a large-scale user study of 10,000 data points to evaluate user preferences towards difference settings, i.e., calculate $P(y | do(x))$ in the Structual Causal Model, where $x$ is a high-level facial attribute.
}
\label{fig:example}
\end{figure}

Powered by Amazon Mechanical Turk (AMT),
we perform a large-scale user study of 10,000 data points to evaluate user preferences towards difference settings, i.e., calculate $P(y | do(x))$ in the Structural Causal Model, where $x$ is a high-level facial attribute.
In this study, $x$ has five options: \texttt{Female}, \texttt{Heavy Makeup}, \texttt{Lipstick}, \texttt{Big Nose} and \texttt{Aged}.
We pick $50$ celebrities from the CelebA dataset, and for each celebrity, we let the users to evaluate the original image and five translated images.
Figure~\ref{fig:example} illustrates experimental results, showing a descending order of preference level,
which aligns with existing psychologist studies~\cite{rhodes2000sex, little2002partnership}.

\section{Conclusion}\label{sec:conclusion}
In this work, we discussed recent advances in photo-realistic image editing and their applications in causal discovery. We introduced the beauty learning problem as an example, showing that beauty exists, is learnable by machines, and can be manipulated by changing latent representations.
Accordingly, we leverage a large-scale user study to counterfactually investigate causal effects from facial semantics to beauty enhancements.
We anticipate more studies in the future to utilize the proposed framework as a powerful tool for causal discovery and beyond.

\section*{Acknowledgement}
This extended abstract is based on~\cite{liu2019understanding,liu2019mining}. The author thanks colleagues at ObEN for their wonderful works.

{
\small
% \footnotesize
\bibliographystyle{IEEEtran}
\bibliography{db}
}

\end{document}